%% file: main_arxiv.tex
\documentclass[10pt,twocolumn,letterpaper]{article}

\usepackage[pagenumbers]{cvpr} 

\usepackage{graphicx}
\usepackage{amsmath}
\usepackage{amssymb}
\usepackage{booktabs}
\usepackage{tabu}

\input{math_definition.tex}
\newcommand\mypara[1]{\vspace{0.5mm}\noindent\textbf{#1}}

\newcommand{\ourmethod}{{OPS}\xspace}

\newcommand{\apf}{$\text{AP}_{50}$}

\usepackage[pagebackref,breaklinks,colorlinks]{hyperref}

\usepackage[capitalize]{cleveref}
\crefname{section}{Sec.}{Secs.}
\Crefname{section}{Section}{Sections}
\Crefname{table}{Table}{Tables}
\crefname{table}{Tab.}{Tabs.}

\def\cvprPaperID{12289} 
\def\confName{CVPR}
\def\confYear{2023}

\begin{document}


\title{Towards Open-World Segmentation of Parts}


\author{Tai-Yu Pan$^{1,}$\thanks{This work was done during an internship at Adobe Research.}
\qquad
Qing Liu$^2$
\qquad
Wei-Lun Chao$^1$
\qquad
Brian Price$^2$ \\
$^1$The Ohio State University \qquad $^2$Adobe Research \\
\texttt{\small \{pan.667, chao.209\}@osu.edu \qquad \{qingl, bprice\}@adobe.com}
\\
{\color{magenta} \texttt{\small https://github.com/tydpan/OpenPartSeg}}
}
\maketitle

\input{main/abs.tex}
\input{main/introduction.tex}

\input{main/related.tex}

\input{main/approach.tex}
\input{main/exp.tex}

\input{main/disc.tex}

{\small
\bibliographystyle{ieee_fullname}
\bibliography{main}
}

\newpage
\appendix
\input{arxiv/suppl}

\end{document}

%% file: math_definition.tex
\usepackage{amssymb}
\usepackage{amsmath,amsfonts}
\usepackage{amsopn}
\usepackage{bm} 
\usepackage{multirow}
\usepackage{tabularx}

\newcommand{\field}[1]{\mathbb{#1}}
\newcommand{\R}{\field{R}} 

\newcommand{\ProbOpr}[1]{\mathbb{#1}}

\newcommand{\expect}[2]{%
\ifthenelse{\equal{#2}{}}{\ProbOpr{E}_{#1}}
{\ifthenelse{\equal{#1}{}}{\ProbOpr{E}\left[#2\right]}{\ProbOpr{E}_{#1}\left[#2\right]}}} 
\newcommand{\var}[2]{%
\ifthenelse{\equal{#2}{}}{\ProbOpr{VAR}_{#1}}
{\ifthenelse{\equal{#1}{}}{\ProbOpr{VAR}\left[#2\right]}{\ProbOpr{VAR}_{#1}\left[#2\right]}}} 

\newcommand{\sS}{\mathcal{S}}

\newcommand{\sU}{\mathcal{U}}

\newcommand{\eat}[1]{}

%% file: main/abs.tex
\begin{abstract}
Segmenting object parts such as cup handles and animal bodies is important in many real-world applications but requires more annotation effort. The largest dataset nowadays contains merely two hundred object categories, implying the difficulty to scale up part segmentation to an unconstrained setting. To address this, we propose to explore a seemingly simplified but empirically useful and scalable task, \emph{class-agnostic part segmentation}. In this problem, we disregard the part class labels in training and instead treat all of them as a single part class. We argue and demonstrate that models trained without part classes can better localize parts and segment them on objects unseen in training. We then present two further improvements. First, we propose to make the model object-aware, leveraging the fact that parts are ``compositions'', whose extents are bounded by the corresponding objects and whose appearances are by nature not independent but bundled. Second, we introduce a novel approach to improve part segmentation on unseen objects, inspired by an interesting finding --- for unseen objects, the pixel-wise features extracted by the model often reveal high-quality part segments. To this end, we propose a novel self-supervised procedure that iterates between pixel clustering and supervised contrastive learning that pulls pixels closer or pushes them away. Via extensive experiments on PartImageNet and Pascal-Part, we show notable and consistent gains by our approach, essentially a critical step towards open-world part segmentation. 
\end{abstract}  

%% file: main/introduction.tex
\section{Introduction}
\label{sec:intro}

Segmenting ``objects'' from images, such as cup, bird, vehicle, etc., is a fundamental task in computer vision and has experienced a series of breakthroughs in recent years thanks to deep learning~\cite{fcn,maskRCNN,mask2former} and large-scale data~\cite{lin2014microsoft,LVIS,OpenImages}. In many real-world applications like object grasping, behavior analysis, and image editing, however, there is often a need to go beyond ``objects'' and dive deeper into their compositions, \ie, ``parts''; for example, to segment cup handle, bird wing, vehicle wheel, etc.

Arguably, the most straightforward way to tackle this problem is to perform part ``instance'' segmentation, \emph{treating each object part as a separate class; each appearance as a separate instance.} A model then must localize parts, classify them, and demarcate their boundaries. In object-level instance segmentation \cite{maskRCNN}, these three sub-tasks are usually approached simultaneously, or at least, share a model backbone. Such a multi-task nature enables the model to benefit from the complementary cues among sub-tasks to attain higher accuracy. For instance, the shapes of segments often entail the class labels and vice versa. 

Segmenting parts in this way, however, limits their scope to the closed world. That is, the learned model may not, or by default should not\footnote{The need to assign a ``seen-class'' label to every detected segment discourages the model from detecting segments that correspond to ``unseen-class'' classes in the first place.}, generalize to object categories (and their corresponding parts) that are unseen during training. Although the largest dataset nowadays for part segmentation, PartImageNet \cite{he2021partimagenet}, has covered more than a hundred object categories, a scale similar to representative object-level datasets like MSCOCO~\cite{lin2014microsoft} and OpenImages \cite{OpenImages}, it is arguable not enough to cover the need in the wild.
 
To equip the model with the open-world capability --- the ability to segment parts for unseen objects --- we propose to chop off the ``classification'' function from the model, as it is simply not applicable to unseen parts.
Namely, we remove the pre-defined fences among different object parts and instead assign a single part class to them (\ie, class-agnostic). 
At first glance, this design choice may seem like a purely simplified version of the original problem or an unavoidable compromise. 
\emph{However, we argue that it indeed helps improve the model's open-world generalizability.} 

Concretely, in training a model to correctly classify \emph{seen} object parts, we implicitly force the model to classify future unseen object parts into the background, suppressing their chances to be detected and segmented. By treating all the {seen} object parts as a single class, we remove the competition tension among them and in turn encourage the model to pay more attention to differentiating ``parts'' and ``non-parts''. As a result, unseen parts appear to be more like the test data in conventional supervised learning; the model can more likely detect them. Besides, removing the competition tension also encourages the model to learn the general patterns of parts, which can potentially improve the segmentation quality on unseen parts. In~\cref{sec:exp}, we empirically demonstrate the effectiveness of class-agnostic training in segmenting parts from unseen objects. 
 
We propose two further improvements towards open-world \emph{class-agnostic} part segmentation. First, \textbf{we incorporate into the model a unique semantic cue of parts.} Compared to objects which are usually considered as ``entities'', \ie, things that can exist and appear distinctively and independently, object parts are ``compositions'', located within an object and often appearing together in a functionally meaningful way. We hypothesize that by making models aware of this object-part relationship, the resulting segmentation quality can be improved. To this end, we propose to introduce \emph{class-agnostic object masks} (\eg, extracted by an off-the-shelf segmentation model) as an additional channel to the model. While extremely simple, we found this approach highly effective, leading to notable gains, especially on unseen objects. Moreover, it is model-agnostic and can easily be incorporated into any network architecture. 

Second, \textbf{we propose a novel way to fine-tune the model using unlabeled data}, \eg, data it sees in its deployed environment, which may include unseen objects. We found that on unseen objects, pixel-wise features the model internally extracts often reveal high-quality segment boundaries. To take advantage of this, we propose a self-supervised approach to adapt the model backbone, which iterates between online pixel clustering (\eg, using k-means) and supervised contrastive learning using the cluster assignment. Concretely, we update the model backbone to pull pixels of the same clusters closer; push pixels between different clusters farther away. As will be demonstrated in~\cref{sec:exp}, this approach leads to a consistent gain on unseen objects and can be further improved via a combination with self-training~\cite{lee2013pseudo,chen2021gradual}. Please see~\cref{fig:pipeline} for an illustration.

We validate our proposed approach, which we name Open Part Segmenter (\ourmethod), on two part segmentation datasets, PartImageNet \cite{he2021partimagenet} and Pascal-Part~\cite{chen2014detect}. We train the model on PartImageNet, and evaluate it on a PartImageNet out-of-distribution set and Pascal-Part: to our knowledge, we are the first to conduct a cross-dataset study for part segmentation. Data in these two sets contain a variety of unseen objects, and we use class-agnostic Average Precision (AP) as the metric. We show that \ourmethod achieves significant and consistent gains against the baselines. On PartImageNet, we improve the AP from $38.21$ to $42.61$; on Pascal-Part, we improve from $9.48$ to $23.02$, almost a $142.8\%$ relative gain. Importantly, all our proposed components --- \emph{class-agnostic segmentation, object mask channel, and self-supervised fine-tuning} --- contribute to the gain. Moreover, 
if given ground-truth object masks (\eg, form a user in an interactive setting), \ourmethod can encouragingly improve the AP to $85.12$ on PartImageNet and $25.26$ on Pascal-Part, making it a highly flexible approach.
Our analyses further reveal cases that \ourmethod can segment even finer-grained parts than the ground truths, essentially a critical step towards open-world part segmentation.

%% file: main/related.tex
\section{Related Work}
\label{sec:related}

\begin{figure*}[t]
    \centerline{\includegraphics[width=0.95\linewidth]{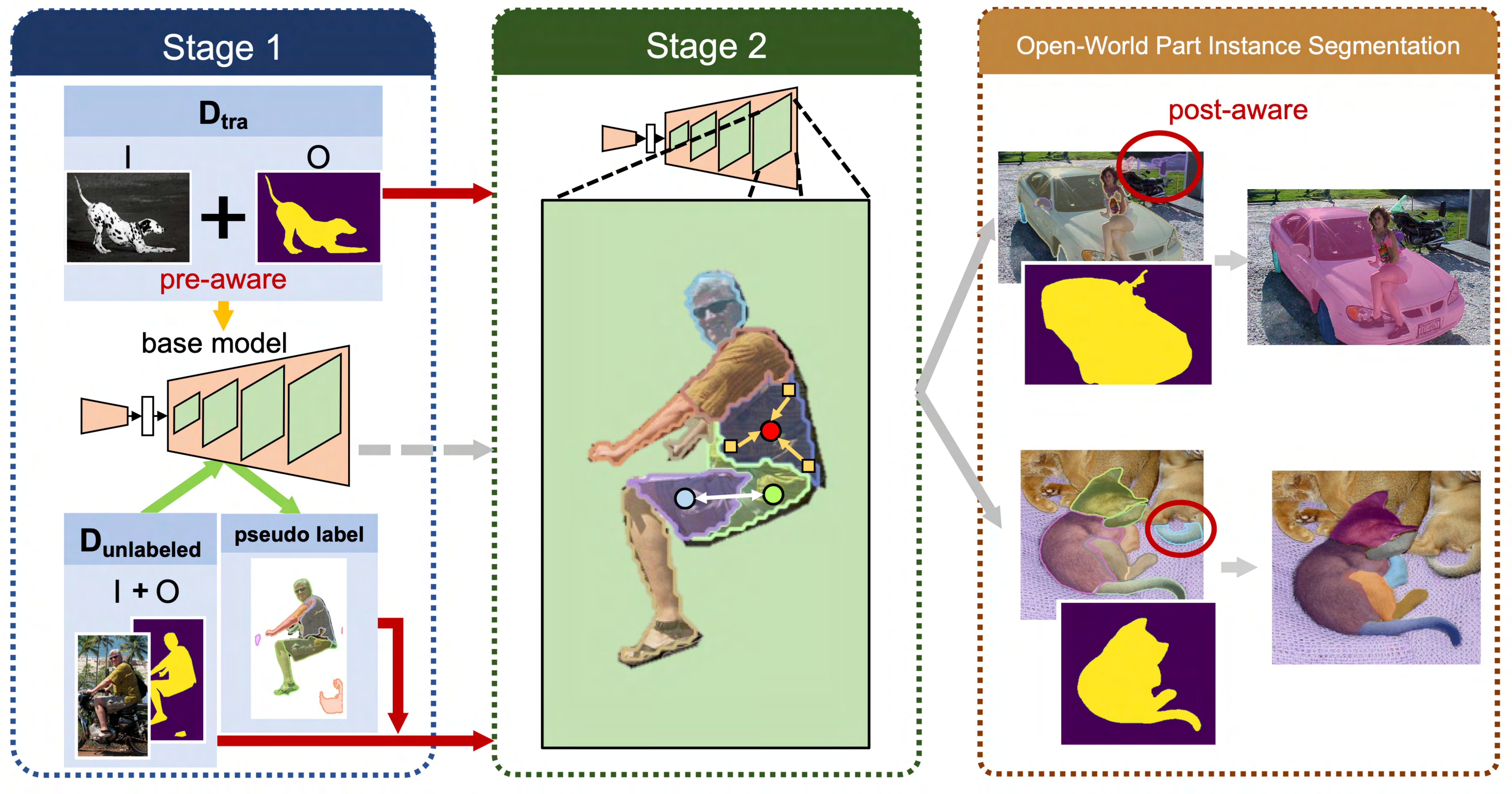}}
    \vspace{-2mm}
    \caption{\small \textbf{Illustration of our pipeline.} We claim that class-agnostic training is the key to the open-world part instance segmentation. In stage 1, we train the model with labeled data treating all part categories as a single class. We then learn with the unlabeled data in stage 2 with our novel fine-tuning approach: (1) self-training (ST): we use pseudo labels generated for unlabeled data by the base model and jointly learn with (2) self-supervised (SS): we discover pseudo parts on the feature map and learn the affinity within a part and contrast between different parts. We also propose object-aware learning including pre-aware and post-aware, which improves the segmentation quality for parts, especially of unseen objects.}
    \label{fig:pipeline}
    \vspace*{-4mm}
\end{figure*}

\mypara{Image segmentation}
aims to divide pixels into semantically meaningful groups. Semantic segmentation~\cite{fcn,pyramid,atrous,zhou2022rethinking} tackles this problem by classifying each pixel into a predefined semantic class, ignoring the fact that pixels of the same classes may belong to different instances. Instance segmentation~\cite{maskRCNN,lee2020centermask,chen2019hybrid,cheng2022sparse}, in contrast, aims to group pixels by semantic instances (\eg, different bird individuals), and at the same time assigns each instance a class label. Panoptic segmentation~\cite{krillov2019panoptic,li2021panopticfcn} further combines instance segmentation of ``things'' (\eg, cars, birds) and semantic segmentation of ``stuffs'' (\eg, sky, grass) into a unified problem. 

Most image segmentation works focus on ``objects'' as the basic class labels~\cite{maskRCNN}. Relatively fewer works treat ``parts'' as the basic class labels~\cite{he2021partimagenet,chen2014detect,cgpart,float,Tritrong,Saha}. In these works, part segmentation is mostly solved as a semantic segmentation problem, despite the fact that an object may contain multiple part instances of the same class (\eg, wheels of a truck). In applications like image editing and object grasping, it is often more important to localize part instances that users/robots can directly interact with. In this paper, we therefore focus on part instance segmentation. 

Since parts are essentially ``compositions'' of objects, there exists a natural (hierarchical) relationship between them. Several works have attempted to leverage this information, mostly in solving hierarchical segmentation over objects and parts together and requiring specifically developed model architectures~\cite{degeus2021panopticparts,Panoptic-PartFormer}. In our work, we focus on parts, assuming that we can obtain object-level masks from an off-the-shelf detector. Without developing a new model architecture, we propose two fairly simple, model-agnostic ways to leverage the object-part relationship. Our approach is particularly suitable for an interactive task like image editing since the model can directly take input from users to generate ``targeted'' part segmentation results.

\mypara{Open-world recognition.}
The visual recognition community has gradually moved from recognizing a pre-defined set of classes to tackling classes that are not seen in the training phases (\eg, rare animals, newly produced products, etc.), \eg, zero-shot learning~\cite{romera2015embarrassingly, xian2017zero, han2021contrastive}. Many recent works in an open-world setting aim to ``classify'' these unseen classes, for example, by exploiting external knowledge such as word vectors~\cite{mikolov2013efficient}, CLIP embedding~\cite{radford2021learning}, etc.

Another approach, which we follow, is to ``localize'' these unseen items while ignoring the need to classify them~\cite{li2017instance,fan2019s4net,qi2021open}. This approach is particularly useful in image editing, manipulation, etc. The closest work to ours is~\cite{qi2021open}, but there are distinct differences. First, they focused on objects while we focus on parts. Second, we propose unique improvements for part segmentation, building upon the insights that parts are ``compositions'', not ``entities'', while they focus on object ``entities''. Further, we propose a novel way to fine-tune our model with unlabeled data.

\mypara{Semi-supervised learning, domain adaptation, and test-time training.}
Our fine-tuning approach on unseen objects is reminiscent of the tasks of semi-supervised learning~\cite{papandreou2015weakly,kervadec2019curriculum,ouali2020semi,huo2021atso}, domain adaptation~\cite{zou2018unsupervised,li2019bidirectional,zhao2019multi,zhang2019category,li2020content,liu2021source,liu2022learning}, and test-time training~\cite{karani2021test,zhu2021test}. However, our fundamental problem setup is different: they mostly assume the unlabeled and labeled data share the same label space; here, we consider the case that our unlabeled data may contain unseen objects or parts.
Nevertheless, we demonstrate that self-training~\cite{lee2013pseudo,zou2018unsupervised,chen2021gradual}, a strong approach in semi-supervised learning and domain adaptation performs favorably in our problem. On top of it, we further propose a novel method taking insights from the detailed investigation of intermediate features within the model for self-supervision on unlabeled data.

%% file: main/approach.tex
\section{Approach}
\label{sec:approach}

\subsection{Class-Agnostic Part Segmentation}
\label{sec:agnostic}


We consider part instance segmentation. We assume that every image contains one salient object and the goal is to segment that object into parts. That is, for each image, $I \in \R^{H \times W \times 3}$, we want to identify and segment the parts of the salient object, $\{(y_n, m_n)\}_{n=1}^N$, where $m \in \{0, 1\}^{H \times W}$ is the binary mask of a part and $y$ is its corresponding part class, assuming the object contains $N$ parts (some of whom may share the same class label). Let us denote by $\sS$ the set of object classes seen in the training data $D_{tr}$.

In a close-world setting, images in the test data $D_{te}$ are expected to only contain objects that belong to $\sS$. In the open-world setting we target, this constraint is removed, allowing the test data 
to contain objects unseen in $D_{tr}$. We denote the set of those unseen object classes by $\sU$.

Since the test data $D_{te}$ may contain unseen objects, their corresponding part labels are, by default, not fully covered by $D_{tr}$. To address this problem, we define the open-world part segmentation problem in a class-agnostic way. That is, we treat all the parts in the test data $D_{te}$ as a single class $y_n = 1$.
The goal is therefore to localize and segment object parts without assigning them class labels.





\mypara{Class-aware v.s. class-agnostic training.} Even though we define the evaluation protocol to be class-agnostic, it may still be beneficial to incorporate part classes during training. To investigate the underlying effect, we compare training a model in a class-agnostic way (\ie, $y_n = 1$) and a class-specific way (\ie, using the original $y_n$ in $D_{tr}$). In evaluation, we simply replace the part class labels predicted by the class-specific model with $1$.

We leave the detailed experimental setup and results in \cref{sec:exp}. In short, in an open-world setting, we observe a consistent improvement by training the model in a class-agnostic way. 
Specifically, class-agnostic training enables the model to localize unseen parts with a higher recall and segment them more accurately. 

In the rest of this paper, we will then employ class-agnostic training to learn the part segmentation model.

\subsection{Object-Aware Learning and Inference}
\label{sec:alpha}

Object parts, by definition, are the ``compositions'' of the corresponding object. That is, the extent of them should not go beyond the extent of the object; object parts belonging to the same object instance should appear closely and be detected within the same object mask.
While a conventional instance segmentation model may learn such prior knowledge from the training data, we argue that directly incorporating it into the model's prediction is a more straightforward way to take advantage of such a strong cue.
 
In this paper, we propose two simple yet highly effective ways to incorporate object masks into part segmentation.


\mypara{Post-aware.} The first way is post-processing. That is, after the model already outputs part segments, we directly remove all the portions outside the given object mask. 


\mypara{Pre-aware.} We argue that the object-awareness should also be proactively included before making predictions. Instead of post-processing afterward, we hypothesize that the model can learn to incorporate the object-part relationship to directly generate higher-quality part segments. More specifically, we incorporate the class-agnostic object mask as an additional channel to the input image $I$, \ie appending $I$ with $O \in \{0, 1\}^{H \times W}$ to become $I' \in \R^{H \times W \times 4}$.


Another potential benefit from this approach is that it makes predicting the background also object-aware, which could potentially improve the recall of part segmentation, especially for unseen objects. Normally, a model predicts background by $P(y=BG|I)$. When the training data are not labeled comprehensively and contain unlabeled objects, the model is forced to predict them as background, making its prediction more conservative. By incorporating the object mask as an input, the model is actually learning $P(y=BG|I, O)$ for each pixel. That is, it can rely on the input $O$ to determine the background region and focuses more on how to segment parts within $O$.




\mypara{Sources of object masks.} There are multiple ways to obtain object masks in practice. Thanks to the large-scale datasets for object-level instance segmentation, such as MSCOCO~\cite{lin2014microsoft}, LVIS~\cite{LVIS}, and OpenImages\cite{OpenImages}, we can train an object instance segmentation model to detect more classes than existing part segmentation datasets contain.

Moreover, considering the application of image editing in an \emph{interactive} environment, users can always define the object region of interest and even refine its boundary. All of the above suggests that the object-awareness is the feasible solution for part segmentation. 

In this paper, we investigate both scenarios. We call the object masks obtained by an object detector ``imperfect'' masks. We assume that the users are able to provide accurate (almost ``perfect'') object masks and simulate such a scenario by using the ``ground-truth'' masks provided by the dataset. In short summary, we obtain huge improvement with both. We leave more details in \cref{sec:exp}.





\begin{figure}[t]
    \centerline{\includegraphics[width=0.9\linewidth]{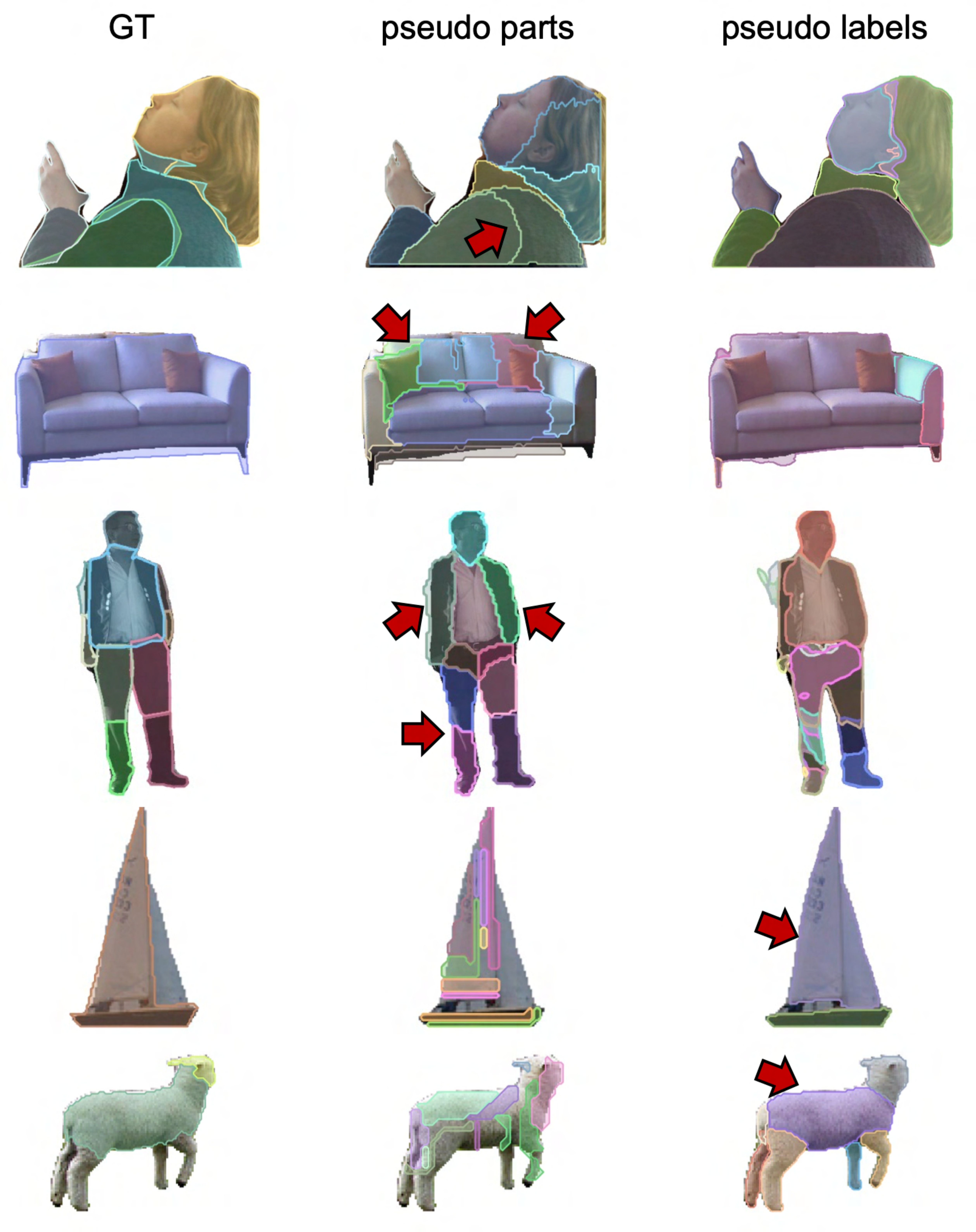}}
    \vspace*{-2mm}
    \caption{\small \textbf{Comparison between pseudo parts for self-supervised (SS) and pseudo labels for self-training (ST) of our fine-tuning approach.} In this figure we use red arrow to mark better parts. When pseudo labels are bad (row 3), we find that pseudo parts on the feature map can actually reveal good ones. They are complementary to each other in \ourmethod.}
    \label{fig:labels}
    \vspace*{-5mm}
\end{figure}

\subsection{Learning with Unlabeled Data}
\label{sec:learning}

To achieve the goal of part instance segmentation in the wild, learning with unlabeled data is a more promising route than waiting for more data to be human-annotated. In this section, we investigate and present approaches to improve the part segmentation model on unlabeled data. In practice, these unlabeled data may come from existing object-level datasets (but without the part annotations), from the web, or even from the deployed environment of the model. In our experiments, we focus on the last scenario. We assume that we have access to the images of the test data $D_{te}$ but without any labels. 

The first way we investigate is self-training (ST). That is, we generate the pseudo labels on $D_{te}$, using the model trained on the labeled training set $D_{tr}$. We then fine-tune the model using the pseudo-labeled data from $D_{te}$ and labeled data from $D_{tr}$ jointly. In our experiments, we observe notable gains with this approach, especially on unseen objects.
However, we also find that the pseudo labels in some cases are already inferior, for example, the human in the third row of \cref{fig:labels}. Fine-tuning with such pseudo-labels can hardly improve these cases or sometimes even have negative effects on the model. 






\mypara{Pseudo parts by pixel-level clustering.} We dig deeper into those inferior cases. We found that the majority of errors come from either under-segmentation, \eg, the model segments the whole object, or boundary mis-localization. We surmise that the model itself just cannot recognize the true segment boundaries for these unseen objects.

To verify this hypothesis, we perform k-means clustering on the feature map. Specifically, we look at the feature map right before the model's prediction, which has the pixel-level resolution, and treat the feature vector on each pixel location as a data point. 

To our surprise, for those inferior cases, k-means often clusters pixels into object parts more accurately than the model's prediction. In some cases, the discovered parts are even finer-grained than the ground truths, for example, the sofa and boat in \cref{fig:labels}.  In other words, by comparing pixels to pixels, the features already capture the \emph{relative} discriminative information among pixels to group them into parts. Such information, however, may not be strong enough in an \emph{absolute} sense to trigger the model to produce parts.

With this evidence, we propose a novel self-supervised (SS) fine-tuning approach to strengthen the discriminative information among pixels. Given an unlabeled image, our approach starts with k-means on top of the feature map, followed by a supervised-contrastive-style loss that pulls features of the same cluster closer and pushes features among different clusters farther away.





\mypara{Contrast between parts.} Intuitively we want the features from different clusters to be dissimilar. We use a centroid to represent each cluster and maximize the pair-wise distance between centroids. Formally, 
\begin{equation}
    \mathcal{L}_{c} = \frac{1}{K}\sum_{i, j}^{K}\exp(\frac{-||c^i - c^j||_2^2}{\tau_c}), 
\end{equation}
where $c^i$ and $c^j$ is the $i$-th and $j$-th centroid from the $K$ clusters; $\tau_c$ is the temperature. 

\mypara{Affinity within parts.} In addition, we want the features within the same cluster to be similar with the following affinitative loss, 
\begin{equation}
    \mathcal{L}_{a} = \frac{1}{N}\frac{1}{K}\sum_i^K\sum_m^N\exp(\frac{||c^i - p^i_m||^2_2}{\tau_a}), 
\end{equation}
where $p^i_m$ is the $m$-th pixel feature within $i$-th cluster; $\tau_a$ is the temperature.

\mypara{Overall loss.} The overall loss for self-supervised fine-tuning is a combination of contrastive and affinitative loss: $\mathcal{L}_{SS} = \lambda_c\mathcal{L}_c + \lambda_a\mathcal{L}_a$, where $\lambda_c$ and $\lambda_a$ are the weights for balancing two losses. 



\subsection{Open Part Segmenter}
Finally, we combine our proposed methods in \cref{sec:agnostic}, \cref{sec:alpha}, and \cref{sec:learning} into a single pipeline named Open Part Segmenter (\ourmethod), aiming to achieve open-world part instance segmentation (cf.~\autoref{fig:pipeline}). We first train the base part segmentation model in a class-agnostic way on labeled data $D_{tr}$, with object masks. We then perform inference with the base model on the unlabeled data $D_{te}$ to obtain the pseudo labels (predicted by the model directly) and pseudo parts (via clustering on the features). We then fine-tune the base model with (i) the contrastive and affinitative losses and (ii) the supervised loss using the (pseudo) labels on both $D_{tr}$ and $D_{te}$. The proposed pipeline is simple but effective, showing dramatic improvements and strong generalizability to unseen objects and parts. See \cref{sec:exp} for details.

We note that in theory we can perform more rounds of self-training to improve the results and we investigate such ideas in the supplementary material. We see consistent improvement but eventually, we see a diminishing return.

%% file: main/exp.tex
\section{Experiments}
\label{sec:exp}

\subsection{Setup}
\label{sec:setup}
This paper aims to achieve open-world part instance segmentation. To validate the concept of our problem and the effectiveness of our approaches, we use the following datasets. We assume that we have a base labeled dataset and access to unlabeled data in the model's deployed environment for training. 

\mypara{PartImageNet\cite{he2021partimagenet}.} This dataset contains $16540$ / $2957$ / $4598$ images and $109$ / $19$ / $30$ object categories in train / val / test split which provide high-quality part segments. The dataset already considers the out-of-distribution (OOD) setting so the object categories in these three subsets are non-overlapped. We also hold out a subset of data from train with a similar size as val to tune the hyper-parameters. 

\mypara{Pascal-Part\cite{chen2014detect}.} To further extend our study to a more realistic open-world setting for part segmentation, we also use Pascal-Part for cross-dataset evaluation. This dataset is a multi-object multi-part benchmark that contains $4998$ / $5105$ images for train / val, covering 20 object categories. Following Singh \etal~\cite{float}, we parse the annotations into Pascal-Part-58/108, which the number here indicates the number of part categories. To serve our purpose, we convert the labels to instance annotations and crop the objects from scene images to become single-object multi-part as PartImageNet~\cite{he2021partimagenet}.

\mypara{Evaluation.} We adopt standard evaluation metrics for instance segmentation, average precision AP and \apf~\cite{lin2014microsoft}. To measure the segmentation quality of predicted parts on OOD and different datasets, we evaluate them in a class-agnostic way. In other words, we treat all the parts as a single class during evaluation regardless of whether the model was trained with or without part classes. 

\subsection{Implementation}
\mypara{Part Segmentation.} We apply Mask2Former~\cite{mask2former} for our part instance segmentation model. The backbone is ResNet-50~\cite{he2016deep} pre-trained on ImageNet-1K~\cite{ILSVRC15}. The input image size is $512$ and by default, it applies large-scale jittering~\cite{ghiasi2021simple} with the minimum and maximum scale of $0.1$ and $2.0$ respectively. In testing time, the shortest edge of the image is resized to $[400, 666]$. We train the base model on PartImageNet~\cite{he2021partimagenet} with batch size $16$ for $90$K iterations. The learning rate starts at $0.0001$ and is decreased by $10$ at $60$K and $80$K. All the experiments are conducted on $8$ Nvidia A100. 

We compare the class-aware and class-agnostic training. The former uses the part classes as annotated while the latter treats all part classes as a single class. Both models are evaluated in a class-agnostic way for a fair comparison. 

\mypara{Object mask for object-awareness.} We demonstrate that the part predictions should be object-aware and investigate two approaches: post-aware and pre-aware. This first approach is a post-processing method that removes all predictions that are outside the object masks. In the second approach, we concatenate the object mask as an additional channel to the input image and hypothesize the model can learn such a relationship between parts and objects.

As mentioned in \cref{sec:alpha}, we have multiple ways to access the object masks. In the following experiments, we obtain the imperfect object masks by taking the trained model on MSCOCO~\cite{lin2014microsoft} from \cite{mask2former} and performing inference on our data. We also show upper-bound results of using perfect object masks. 

\mypara{Learning with unlabeled data.} We fine-tune the base model trained on PartImageNet~\cite{he2021partimagenet} train data with our novel self-supervised (SS) + self-training (ST) approach on unlabeled data. We consider two settings: (1) fine-tuning on PartImageNet val and evaluate on test and (2) fine-tuning on Pascal-Part~\cite{chen2014detect} train and evaluate on Pascal-Part val. Both measure the generalizability to the OOD data and the second is even cross-set. 

Each training batch of 32 is an even mix of labeled/pseudo-labeled data and unlabeled pseudo parts data.
For pseudo labels in ST, we use predictions from the base model that have confidence scores larger than $0.1$. For pseudo parts in SS, we use $K=10$ for online K-Means on normalized features $\mathcal{F}'$. We use $1$ for $\tau_c$ and $\tau_a$, $10$ for $\lambda_c$ and $0.5$ for $\lambda_a$. We fine-tune the model for $30$K and $10$K iterations with learning rate $1\mathrm{e}{-6}$ with imperfect and perfect object masks respectively. 

\subsection{Main Results on PartImageNet}
\label{sec:part}
In this section, we show our results on PartImageNet~\cite{he2021partimagenet} val and test set. Both sets are OOD from the train set. Please see more information in~\cref{sec:setup}. We follow the rationale of our proposed methods step by step to provide a comprehensive study with the empirical results on the open-world part instance segmentation problem. 

\mypara{Class-aware v.s. class-agnostic.} The former trains the base model with part classes while the latter treats all the part classes as a single class. In order to investigate the underlying impact of class labels on the quality of part instance segmentation, we evaluate both approaches in a class-agnostic way. At first glance, we may think the model can leverage more information from part categories in class-aware training to refine the predicted part masks accordingly. However, we find that the class-agnostic training obtains comparable or slightly better performance. As shown in \cref{tab:ag}, $40.01$ from class-agnostic is already better than $39.71$ from class-aware in the plain setting without any other proposed method. The observation still holds when we further include the object masks. This suggests that learning the context alone can already achieve high-quality part segmentation. We hypothesize the model learns more general representation about parts and thus performs well on OOD data. It encourages our proposed method toward the open world. In all of the following experiments, we will directly use class-agnostic training unless stated otherwise. 

\begin{table}
\centering 
\tabcolsep 5pt
\caption{\small \textbf{Comparison between class-agnostic and class-aware training.} We first study the impact of two on PartImageNet~\cite{he2021partimagenet} val set to validate our problem setting and approach. Class-agnostic consistently performs better than class-aware, with (imperfect, perfect) or without (none) object masks.}
\vspace*{-2mm}
\label{tab:ag}
\begin{tabu}{p{1.1cm}|cc|cc|cc}
             & \multicolumn{2}{c|}{none} & \multicolumn{2}{c|}{imperf.} & \multicolumn{2}{c}{perf.} \\
class     & AP          & \apf        & AP            & \apf         & AP          & \apf        \\\tabucline[1.2pt]{-}
agnostic & 40.01       & 70.38       & 41.94         & 73.17        & 85.88       & 96.08       \\
aware    & 39.71       & 69.99       & 41.45         & 72.97        & 84.74       & 95.50      
\end{tabu}
\vspace*{-5mm}
\end{table}



\mypara{Object-aware part segmentation.}
We propose post-aware and pre-aware for object-awareness. While the former uses object masks for post-processing to filter out unreasonable part predictions, the latter includes them as a cue for training. In \cref{tab:newalpha}, we show that both approaches outperform the base model that has no object-awareness (\ie AP $40.01$). It proves the effectiveness of our proposed methods. Furthermore, the two approaches are complementary to each other. Simply combining them obtains the further gain, \ie AP $41.94$ to $45.40$ with imperfect masks and $85.88$ to $87.61$ with perfect masks. 

Here we also see a big jump from not using object masks (AP $40.01$) to using perfect ones (AP $87.61$) in \cref{tab:newalpha}. This is an essential finding: considering image editing in an interactive environment, a user can always refine the object mask until it is satisfying. This encourages the application of our work to the real world. 

Besides the above improvements, our proposed methods are simple and straightforward. It can be easily plugged into most existing algorithms and architectures. 


\begin{table}
\centering
\caption{\small \textbf{Results on post-aware and pre-aware object masks on PartImageNet~\cite{he2021partimagenet} val and Pascal-Part-58~\cite{chen2014detect,float}.} Both approaches outperform the base models that have no object-awareness (none). }
\vspace*{-2mm}
\label{tab:newalpha}
\begin{tabu}{p{2.4cm}cc|cc}
 \multicolumn{1}{l}{} & \multicolumn{2}{c|}{PartImageNet val} & \multicolumn{2}{c}{Pascal-Part-58} \\
 object mask & AP    & \apf  & AP    & \apf \\\tabucline[1.2pt]{-}
 none           & 40.01 & 70.38 & 9.48  & 19.90 \\
\hspace{0.3cm} + post imperf. & 42.48 & 71.03 & 12.44 & 24.40 \\
\hspace{0.3cm} + post perf.   & 47.10 & 75.25 & 13.06 & 24.97 \\\tabucline[0.5pt]{-}\tabucline[0.5pt]{-}
pre imperf.        & 41.94 & 73.17 & 20.27 & 44.24 \\
\hspace{0.3cm} + post imperf. & 45.40 & 74.45 & 23.02 & 47.21 \\\tabucline[0.5pt]{-}\tabucline[0.5pt]{-}
pre perf.          & 85.88 & 96.08 & 25.24 & 45.62 \\
\hspace{0.3cm} + post perf. & 87.61 & 96.19 & 25.26 & 45.67
\end{tabu}
\vspace*{-5mm}
\end{table}

\mypara{Learning with unlabeled data.}
In this section, we demonstrate the effectiveness of our novel fine-tuning approaches, namely self-training (ST) and self-supervised (SS) learning. We fine-tune the base model with PartImageNet~\cite{he2021partimagenet} val and evaluate on the test set. As both val and test are OOD from train, we aim to investigate the generalizability to unseen objects and parts in terms of segmentation quality. 

In \cref{tab:imagenet}, the performance of the base model on val set is AP $41.94$ with imperfect masks. Fine-tuning on val set with SS and ST alone can improve to $42.78$ and $43.12$ respectively. It shows the effectiveness of each individual component. By combining them, our proposed OPS model can get even higher performance. 

Fine-tuning and improving on val set assumes we have access to the unlabeled data without annotations and we can leverage them in an unsupervised learning way. Here we still have another test set OOD from both train and val which is not included in the fine-tuning. In \cref{tab:imagenet}, we also see notable gains with all approaches (AP $40.17$, $40.38$, and $40.43$) compared to the base model (AP $38.96$). It explains that the model learns more generalizability to unseen parts and objects with fine-tuning only on val set. The proposed fine-tuning approach is an important step toward open-world part instance segmentation. 

\begin{table}
\centering
\caption{\small \textbf{Results on PartImageNet~\cite{he2021partimagenet}} We fine-tune the base model on OOD val set with proposed self-supervised (SS) and self-training (ST), with imperfect and perfect object masks. Both outperform the base model.}
\vspace*{-2mm}
\label{tab:imagenet}
\begin{tabu}{p{1.2cm}cc|cc|cc}
 & \multicolumn{1}{l}{} & \multicolumn{1}{l|}{} & \multicolumn{2}{c|}{Val} & \multicolumn{2}{c}{Test} \\
         method  & SS & ST & AP    & \apf  & AP    & \apf  \\\tabucline[1.2pt]{-}
imperf. &&&&& \\
base &     &    & 41.94 & 73.17 & 38.96 & 69.07 \\
              & \checkmark   &    &   42.78    &   74.62    &   40.17    &   70.70    \\
              &     & \checkmark  &   43.12    &    75.03   &   40.38    &    71.10   \\ 
\ourmethod             & \checkmark   & \checkmark  & 43.16 & 74.96 & 40.43 & 71.18 \\ \tabucline[0.5pt]{-}\tabucline[0.5pt]{-}
perf. &&&&& \\
base   &     &    & 85.88 & 96.08 & 83.52 & 94.66 \\
             & \checkmark   &    &   86.09    &  96.35     &  83.81     &    94.94   \\
             &     & \checkmark  &   86.28    &  96.37     &    83.97   &   95.12    \\
\ourmethod             & \checkmark   & \checkmark  & 86.19 & 96.43 & 83.86 & 95.05
\end{tabu}
\vspace*{-5mm}
\end{table}


\begin{figure*}[ht]
    \centerline{\includegraphics[width=1\linewidth]{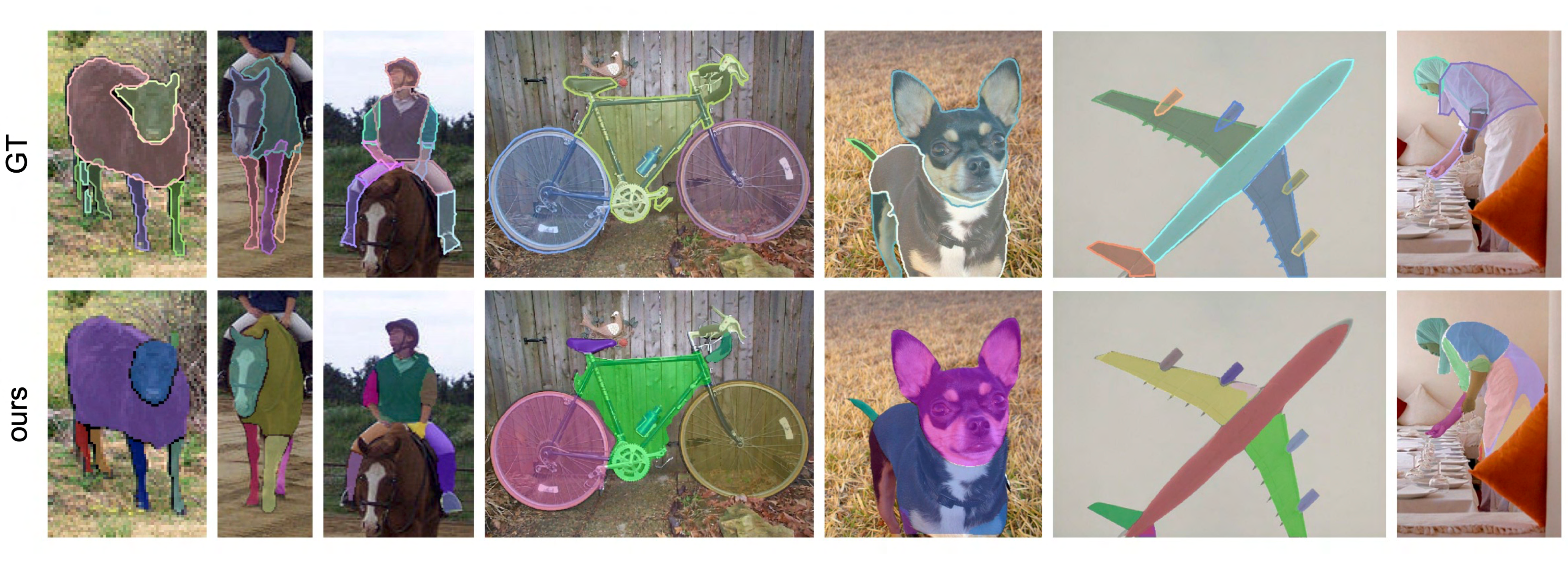}}
    \vspace*{-2mm}
    \vskip-5pt
    \caption{\small \textbf{Visualizations of part segmentation on Pascal-Part-58~\cite{chen2014detect,float}.} The first row is the ground-truth and the second is predictions from our proposed OPS model. Our model only uses part segmentation ground-truth from PartImageNet~\cite{he2021partimagenet}. These high-quality part predictions demonstrate the feasibility of part segmentation in the wild and the effectiveness of our proposed approach.}
    \label{fig:qual}
    \vspace*{-5mm}
\end{figure*}

\subsection{Main Results on Pascal-Part}
\label{sec:pascal}
In \cref{sec:part}, we show notable improvements on PartImageNet~\cite{he2021partimagenet}. Here, we further extend our study to a cross-dataset setting. To this end, we use the same base model trained on PartImageNet train set, and we measure the generalizability of the model to the new test data in Pascal-Part~\cite{chen2014detect}. We evaluate on two sets of ground-truth annotations, Pascal-Part-58 and Pascal-Part-108. We parse the annotations following \cite{float} (see \cref{sec:setup} for more information). To the best of our knowledge, we are the first to investigate cross-dataset scenarios in part instance segmentation. 


\mypara{Object-aware part segmentation.}
We conduct the same empirical study as in \cref{sec:part} on Pascal-Part-58. 
As shown in \cref{tab:newalpha}, the base model performs merely AP $9.48$ and improves to $13.06$ even after proposed post-processing with perfect mask (base + post perf.). With pre-aware imperfect masks, the performance (AP $20.27$) is more than twice as the base model and also outperforms base + post perf. by a large margin. We observe similar results with perfect object masks. Furthermore, the gain is greater than what we have seen in \cref{sec:part}. This suggests the effectiveness of the pre-aware approach, which can recognize more high-quality part segments, especially for data with a larger domain gap. 

\mypara{Learning with unlabeled data.}
In this section, we follow the same route as in \cref{sec:part} but fine-tune the base model using unlabeled data in the Pascal-Part train set. As shown in \cref{tab:pascal}, on Pascal-Part-58, SS improves the base model from AP $20.27$ and $25.24$ to $20.53$ and $27.13$ with imperfect and perfect object masks respectively. With ST, the APs are further boosted to $24.02$ and $27.69$. These relative improvements of $18.4\%$ and $9.7\%$ are much greater than $2.9\%$ and $0.3\%$ in \cref{tab:imagenet} on PartImageNet. 

On the more fine-grained and challenging Pascal-Part-108, we also observe consistent improvements, \ie relative improvements of $11.43\%$ and $22.38\%$ to the base model. 

\mypara{Qualitative results.} In \cref{fig:qual}, we show visualizations of part segmentation using our proposed OPS model on Pascal-Part-58~\cite{chen2014detect,float}. Though our model is only trained using ground-truth annotations from PartImageNet~\cite{he2021partimagenet} in a class-agnostic way, it generates high-quality part segments which align very well with GT.

In short, we demonstrate that our approaches are applicable to not only PartImageNet, OOD but closer domain, but also even more challenging cross-dataset setting. With our approach of learning with unlabeled data, OPS is able to learn better representation for more general parts, which results in superior performance on unseen objects and parts from different data domains.


\begin{table}
\centering
\caption{\small \textbf{Results on Pascal-Part-58 and Pascal-Part-108~\cite{chen2014detect,float}.} We fine-tune the base model on train set with the proposed self-supervised (SS) and self-training (ST), without using any additional ground-truth annotations. We show the performance with imperfect and perfect object masks. All the components in \ourmethod consistently outperform the base model.}
\label{tab:pascal}
\begin{tabu}{p{1.2cm}cc|cc|cc}
 & \multicolumn{1}{l}{} & \multicolumn{1}{l|}{} & \multicolumn{2}{c|}{Part-58 val} & \multicolumn{2}{c}{Part-108 val} \\
    method   & SS & ST & AP    & \apf  & AP    & \apf \\\tabucline[1.2pt]{-}
imperf. &&&&& \\
base &    &    & 20.27 & 44.24 & 16.36 & 30.16 \\
             & \checkmark   &    & 20.53 & 44.91 & 17.95 & 33.03 \\
             &     & \checkmark  & 23.25 & 48.81 & 17.75 & 32.64 \\ 
\ourmethod   & \checkmark   & \checkmark  & 24.02 & 50.10 & 18.23 & 33.72 \\\tabucline[0.5pt]{-}\tabucline[0.5pt]{-}
perf. &&&&& \\
base   &    &    & 25.24 & 45.62 & 13.40 & 29.32 \\
             & \checkmark   &    & 27.13 & 49.08 & 13.75 & 29.76 \\
             &     & \checkmark  & 27.23 & 48.58 & 15.85 & 33.66 \\ 
\ourmethod   & \checkmark & \checkmark & 27.69 & 49.75 & 16.40 & 34.60
\end{tabu}
\vspace*{-5mm}
\end{table}

\subsection{Comparison to Other Baselines}
\label{sec:baselines}
\mypara{Evaluation metric.}
We adopt the standard evaluation metric AP for part ``instance'' segmentation in previous sections. In this section, we evaluate with mIOU, fwIoU, and mACC, which are the standard metrics for semantic segmentation, to compare to other baselines. However, directly applying them without classifying parts will simply merge all the parts and cannot reflect the segmentation quality. To resolve this, we consider an oracle scenario, assigning each segmented part the class label from its highest-overlapped ``ground-truth'' part. 

\begin{table}[t]
\small
\centering 
\tabcolsep 4pt
\caption{\small \textbf{Comparison to baselines on PartImageNet test.} We adopt standard metrics for semantic segmentation to compare with other baselines. \ourmethod outperforms them by a large margin.}
\vspace*{-2mm}
\label{tab:iou}
\begin{tabu} {p{0.7cm}|ccc|ccc}
    & \multicolumn{3}{c|}{imperfect obj.~mask}  & \multicolumn{3}{c}{perfect obj.~mask}     \\
    & mIOU  & fwIoU & mACC & mIOU  & fwIoU & mACC  \\\tabucline[1.2pt]{-}
SLIC & 38.15 & 65.10 & 75.61 & 41.19 & 68.09 & 78.12\\
NC  & 40.80 & 76.87 & 54.99 & 43.11 & 77.44 & 55.17 \\
Fel. & 47.97 & 83.82 & 67.70 & 59.73 & 88.90 & 76.10 \\
OPS & \textbf{64.71} & \textbf{89.41} & \textbf{82.61} & \textbf{91.89} & \textbf{97.97} & \textbf{95.81}
\end{tabu}
\vspace*{-5mm}
\end{table}

\mypara{Results.}
We compare \ourmethod to normalized cut (NC)~\cite{shi2000normalized}, SLIC~\cite{achanta2012slic}, and Felzenszwalb (Fel.)~\cite{felzenszwalb2004efficient} as shown in \cref{tab:iou}. \ourmethod outperforms them by a large margin in all metrics on PartImageNet test. Our method using imperfect masks even outperforms them when they use perfect masks. We argue \ourmethod successfully captures semantic cues to attain high quality. See the supplementary material for more.

%% file: main/disc.tex
\section{Conclusion}
\label{sec:disc}


In this work, we present \ourmethod, a method for part segmentation in an open-world setting. To be robust to unseen parts, we propose \emph{class-agnostic} and \emph{object-aware} learning. Combining with self-training and clustering on unlabeled data, we achieve state-of-the-art on unseen categories.

%% file: arxiv/suppl.tex



\renewcommand{\thetable}{\Alph{table}}
\renewcommand{\thefigure}{\Alph{figure}}
\renewcommand{\theequation}{\Alph{equation}}
\setcounter{table}{0}
\setcounter{figure}{0}
\setcounter{equation}{0}

\clearpage
\appendix
\begin{center}
\textbf{\Large Supplementary Material}
\end{center}


In this supplementary material, we provide details and results omitted in the main text.
\begin{itemize}
    \item \autoref{sec:suppl_reverse}: \textbf{Pascal-Part-58 to PartImageNet.} In contrast with the main paper \cref{sec:pascal}, we further perform the reverse direction of training and evaluation to validate our proposed method.
    \item \autoref{sec:suppl_multi}: \textbf{Multiple-Round Self-Training.} In the main paper \cref{sec:part} we report singe round self-training. In this supplementary, we further explore multiple rounds.
    \item \autoref{sec:suppl_imperfect}: \textbf{Robustness to Imperfect Object Masks} In this section, we visualize more predictions to show the effect of using imperfect object masks in pre-aware setting to support discussions in \cref{sec:alpha} of the main paper.
    \item \autoref{sec:suppl_robustness}: \textbf{Robustness to Unseen Parts.} In this section, we show more cases of applying OPS to unseen objects and unseen parts. Some of our predictions are more reasonable and fine-grained than GT.
    \item \autoref{sec:suppl_base}: \textbf{Comparison to More Baselines.} In this section, we compare to an additional baseline, SCOPS~\cite{hung2019scops}, besides \cref{sec:baselines} in the main paper.
    \item \autoref{sec:suppl_multiobjs}: \textbf{Robustness to Multiple Objects in a Scene. } In this section, we show more cases of applying OPS to an image containing multiple objects. 
     \item \autoref{sec:suppl_qual}: \textbf{More Qualitative Results.}
\end{itemize}

\section{Pascal-Part-58 to PartImageNet}
\label{sec:suppl_reverse}
As claimed in the main paper, our goal is to improve the robustness of the part segmentation model to the unseen parts.
Our proposed method, \ourmethod, utilizes the novel self-supervised (SS) and self-training (ST) fine-tuning approach to learn with unlabeled data.
In Section 4.3 and Section 4.4, we investigate two settings: (1) train the base model on PartImageNet train, fine-tune the base model on PartImageNet val without ground-truth labels, evaluate on both PartImageNet val and PartImageNet test; (2) train the base model on PartImageNet train, fine-tune the base model on Pascal-Part train without ground-truth labels, evaluate on Pascal-Part val. In this supplementary, we further provide the result of using Pascal-Part-58 to train the base model, PartImageNet train set to fine-tune, and PartImageNet val/test set to evaluate. 

In \cref{tab:suppl_imagenet}, we see consistent improvements over the base models with our proposed SS and ST methods. Both SS and ST can improve part segmentation when working alone. By combining SS and ST, our proposed full OPS model achieves further gain, which improves the val set from AP $22.59$ to $25.14$ and the test set from $18.58$ to $20.73$ with imperfect object masks, and val set from AP $36.39$ to $37.93$ and test set from $32.08$ to $33.71$ with perfect object masks. This demonstrates our proposed OPS model indeed achieves improved robustness for part segmentation on unseen objects no matter how it is tested in the cross-dataset setting.

\begin{figure*}[ht]
    \centerline{\includegraphics[width=1\linewidth]{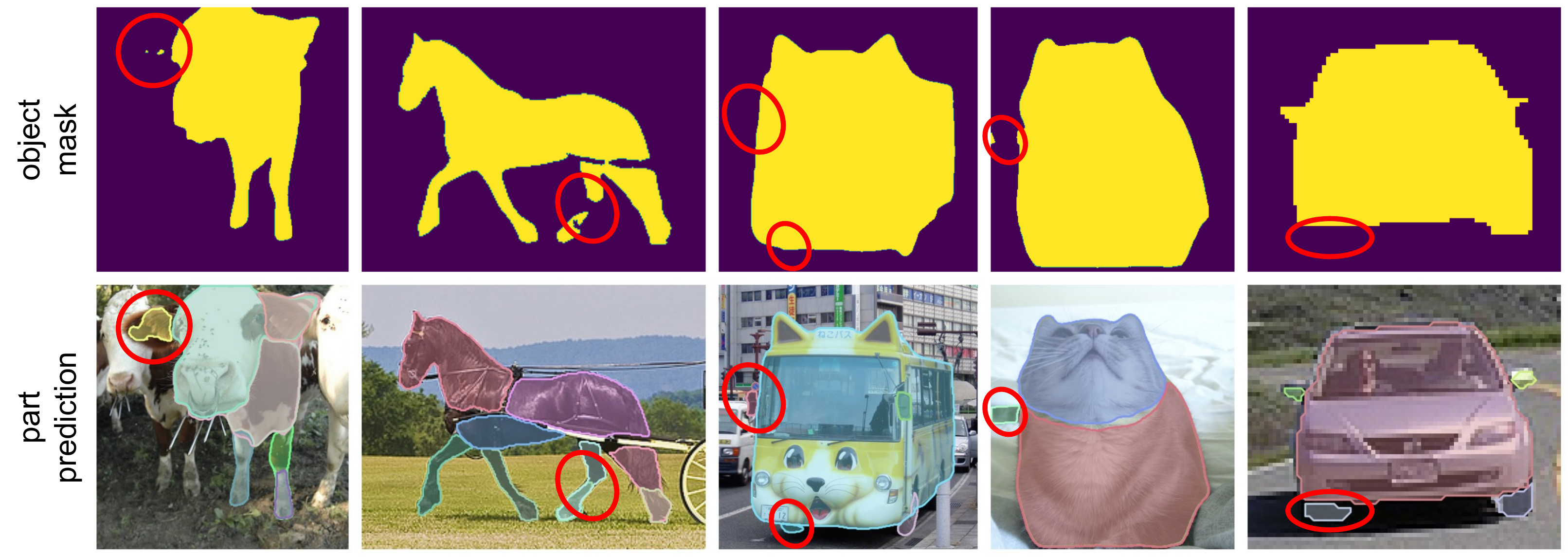}}
    \caption{\small \textbf{Imperfect object masks.} The red circles indicate the parts missed in the imperfect object masks but recovered by OPS part predictions. The model can recognize the shape of the objects and their belonging parts even though it is trained in a class-agnostic way.}
    \label{fig:suppl_alpha}
\end{figure*}

\section{Multiple-Round Self-Training}
\label{sec:suppl_multi}
In the main paper, we report the result of single-round self-training: we use the base model to generate pseudo labels and perform fine-tuning for a single round. In many self-training works, multiple rounds of pseudo-label generation and fine-tuning are usually performed. In this setting, pseudo labels are updated by the fine-tuned model and additional fine-tuning can be applied on top of it. Here, we explore two rounds of self-training for our proposed OPS model, and results are shown in \cref{tab:suppl_multi}.

On PartImageNet, we see the result of OPS gets slightly improved (AP $43.16$ to $43.29$ on the val set and $40.43$ to $40.78$ on test set) with imperfect object masks and the performance is nearly the same with perfect object masks. On Pascal-Part-58, we improve from AP $27.69$ to $27.83$ with perfect object masks, but the performance stays almost the same (AP $24.02$ vs $23.96$) with imperfect object masks. Note that the performance on multi-rounds may not be optimal yet because it requires further mining on pseudo labels, which will be investigated more in our future work.

\begin{table}
\centering
\caption{\small \textbf{Results on PartImageNet~\cite{he2021partimagenet}} We train the base model on Pascal-Part-58~\cite{chen2014detect,float} and fine-tune it on PartImageNet train set with proposed self-supervised (SS) and self-training (ST), with imperfect and perfect object masks. Both outperform the base model.}
\label{tab:suppl_imagenet}
\begin{tabu}{p{1.2cm}cc|cc|cc}
 & \multicolumn{1}{l}{} & \multicolumn{1}{l|}{} & \multicolumn{2}{c|}{Val} & \multicolumn{2}{c}{Test} \\
         method  & SS & ST & AP    & \apf  & AP    & \apf  \\\tabucline[1.2pt]{-}
imperf. &&&&& \\
base &     &    & 22.59 & 48.06 & 18.58 & 39.47 \\
              & \checkmark   &    &   23.60    &   49.90    &   19.27    &   40.64    \\
              &     & \checkmark  &   24.60    &    53.17   &   20.23    &    43.70   \\ 
\ourmethod             & \checkmark   & \checkmark  & 25.14 & 54.18 & 20.73 & 44.46 \\ \tabucline[0.5pt]{-}\tabucline[0.5pt]{-}
perf. &&&&& \\
base   &     &    & 36.39 & 63.01 & 32.08 & 55.24 \\
             & \checkmark   &    &   38.40    &  65.29     &  33.40     &    56.90   \\
             &     & \checkmark  &   36.86    &  66.83     &    32.89   &   59.12    \\
\ourmethod             & \checkmark   & \checkmark  & 37.93 & 67.87 & 33.71 & 59.93
\end{tabu}
\end{table}

\begin{table}
\centering
\caption{\small \textbf{Results of multiple rounds for pseudo labels.} In the main paper, we report the result of a single round, which generates the pseudo labels only once. In the supplementary, we further explore multiple rounds.}
\label{tab:suppl_multi}
\begin{tabu}{lcc|cc}
\multicolumn{1}{c}{}         & \multicolumn{2}{c|}{single round} & \multicolumn{2}{c}{multi rounds} \\
\multicolumn{1}{c}{datasets} & AP   & \apf  & AP  & \apf  \\\tabucline[1.2pt]{-}
PartImageNet val             &               &                 &              &                \\
\hspace{0.6cm}w/ imperf.                   & 43.16         & 74.96           & 43.29        & 74.80          \\
\hspace{0.6cm}w/ perf.                     & 86.19         & 96.43           & 86.18        & 96.41         
\\\tabucline[0.5pt]{-}\tabucline[0.5pt]{-}
PartImageNet test            &               &                 &              &                \\
\hspace{0.6cm}w/ imperf.                   & 40.43         & 71.18           & 40.78        & 71.20          \\
\hspace{0.6cm}w/ perf.                     & 83.86         & 95.05           & 83.86        & 95.05          \\\tabucline[0.5pt]{-}\tabucline[0.5pt]{-}
Pascal-Part-58               &               &                 &              &                \\
\hspace{0.6cm}w/ imperf.                   & 24.02         & 50.10           & 23.96        & 49.80          \\
\hspace{0.6cm}w/ perf.                     & 27.69         & 49.75           & 27.83        & 50.13         
\end{tabu}
\end{table}



\section{Robustness to Imperfect Object Masks}
\label{sec:suppl_imperfect}
In the main paper, we propose to apply object-aware learning which aims to capture the fact that parts are ``compositions'' of their objects. We extract imperfect object masks by an off-the-shelf segmentation model (see main paper for more information) and input with images as an additional channel to RGB. \cref{fig:suppl_alpha} shows that OPS part predictions are able to complete the missing parts even though the imperfect object masks are used for pre-awareness. For example, in the first column, the predictions of the cow recover the right ear region. Similarly, the predictions of the horse in the second column repair the leg. In addition, the model is able to exclude the rein since it is less likely to be a part of the horse. 

\begin{figure*}[ht]
    \centerline{\includegraphics[width=1\linewidth]{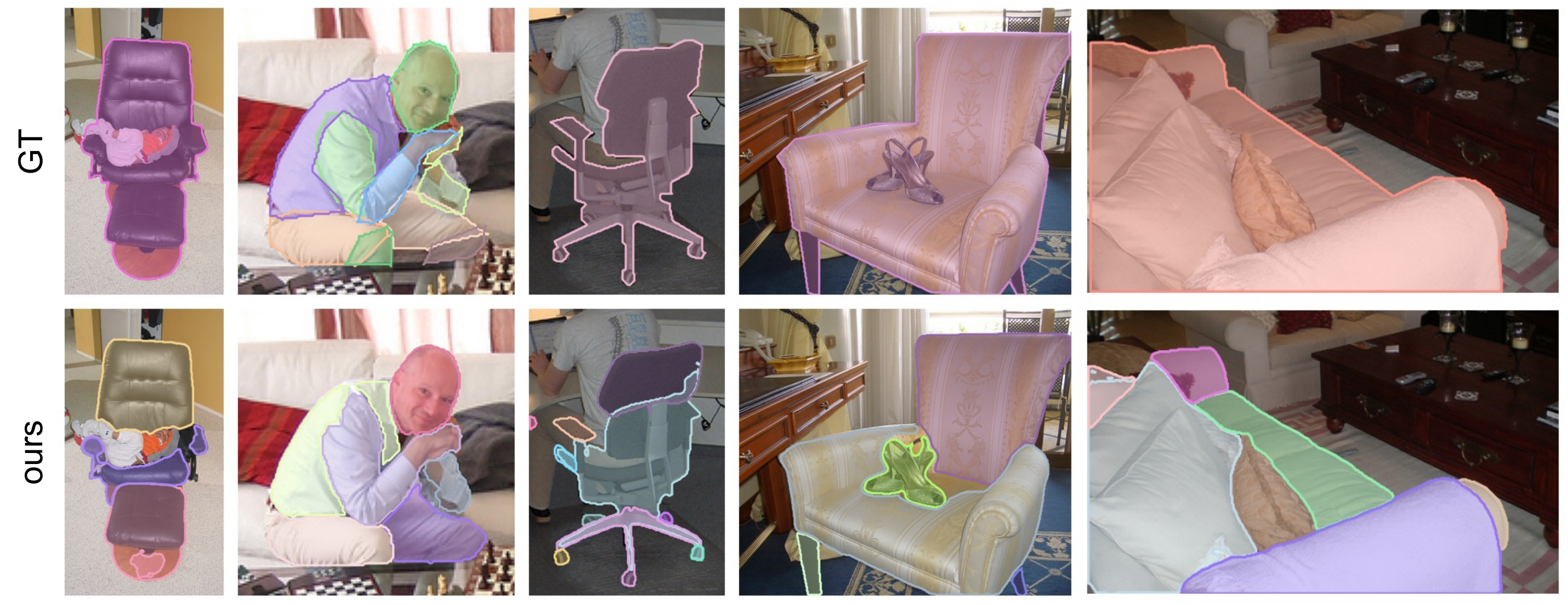}}
    \caption{\small \textbf{Robustness to unseen parts.} OPS shows strong generalizability to objects and parts that are unseen in PartImageNet~\cite{he2021partimagenet}. Furthermore, our method is able to segment parts that are not annotated in Pascal-Part~\cite{chen2014detect}. }
    \label{fig:suppl_qual}
\end{figure*}

\section{Robustness to Unseen Parts}
\label{sec:suppl_robustness}
In \cref{fig:suppl_qual}, we show the robustness of \ourmethod to the unseen objects and parts. Here the test images are from Pascal-Part-58 val, while our OPS model is trained on PartImageNet train and fine-tuned on Pascal-Part-58 train. Some of the predicted parts are not annotated in Pascal-Part~\cite{chen2014detect}. In the first column, the prediction excludes the baby on the chair while the GT labels the whole as a single part. In the third column, the prediction not only finds out the headrest, body, and bottom part of the chair but also discovers all wheels and the armrest. In these cases, our predictions won't get any reward in terms of evaluation metric, e.g., AP or \apf, and they will even get lower scores since the predicted parts are not annotated in the ground-truth. We will explore better evaluation metrics for unlabeled part discovery in our future work.

\section{Comparison to More Baselines}
\label{sec:suppl_base}
\begin{figure}[t]
  \centering
  \includegraphics[width=1\linewidth]{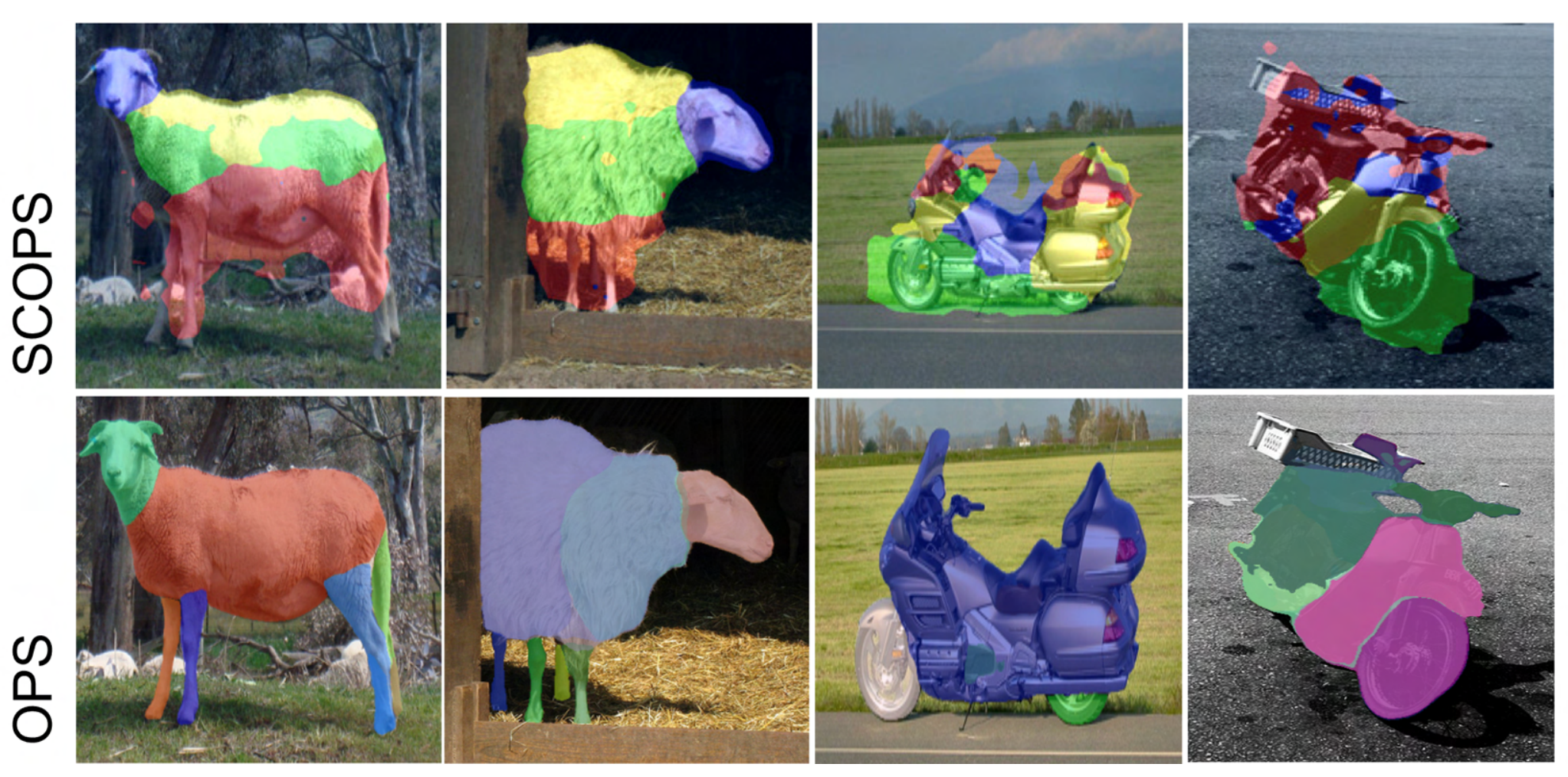}
   \caption{\small \textbf{Comparison to SCOPS~\cite{hung2019scops}} (The results in the first row are directly copied from its paper). \ourmethod has a higher quality of parts.}
   \label{fig:scops}
\end{figure}

In this section, we try to compare \ourmethod to SCOPS~\cite{hung2019scops} as an additional baseline. We note that SCOPS~\cite{hung2019scops} experimented with PASCAL-Part but did not release the checkpoint; it reported object-level IoU (by aggregating parts) but not part-level IoU or AP. Therefore, we perform a qualitative comparison by applying \ourmethod on the PASCAL-Part images that SCOPS~\cite{hung2019scops} showed in their paper. As shown in \cref{fig:scops}, \ourmethod generally leads to a higher quality of parts.

\section{Robustness to Multiple Objects in a Scene}
\label{sec:suppl_multiobjs}
\begin{figure}[t]
  \centering
  \includegraphics[width=1\linewidth]{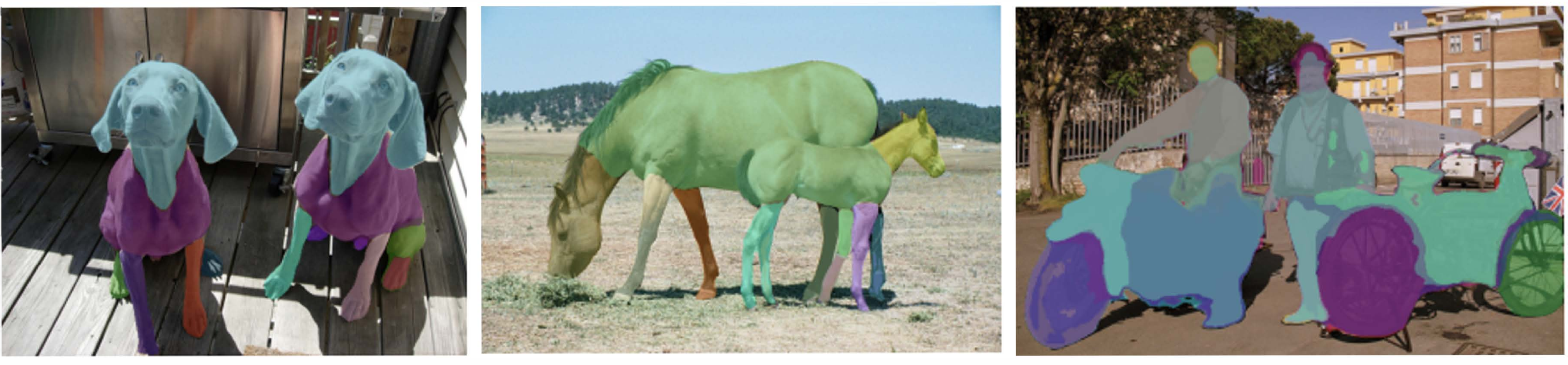}
   \caption{\small \textbf{Qualitative results on multiple objects in an image. }}
   \label{fig:multiobjs}
\end{figure}

We use single-object images for simplicity by following SCOPS~\cite{hung2019scops}.
Meanwhile, our \ourmethod can be applied to a multi-object image \emph{in one pass} by simply including a multi-object mask. 
As shown in \cref{fig:multiobjs}, although in the rightmost case, all object masks are connected without differentiation in the mask channel, \ourmethod still correctly recognizes the parts of each object. In addition, simple post-processing with object masks can further refine the part predictions. 

\section{More Qualitative Results}
\label{sec:suppl_qual}
\begin{figure*}[ht]
    \centerline{\includegraphics[width=1\linewidth]{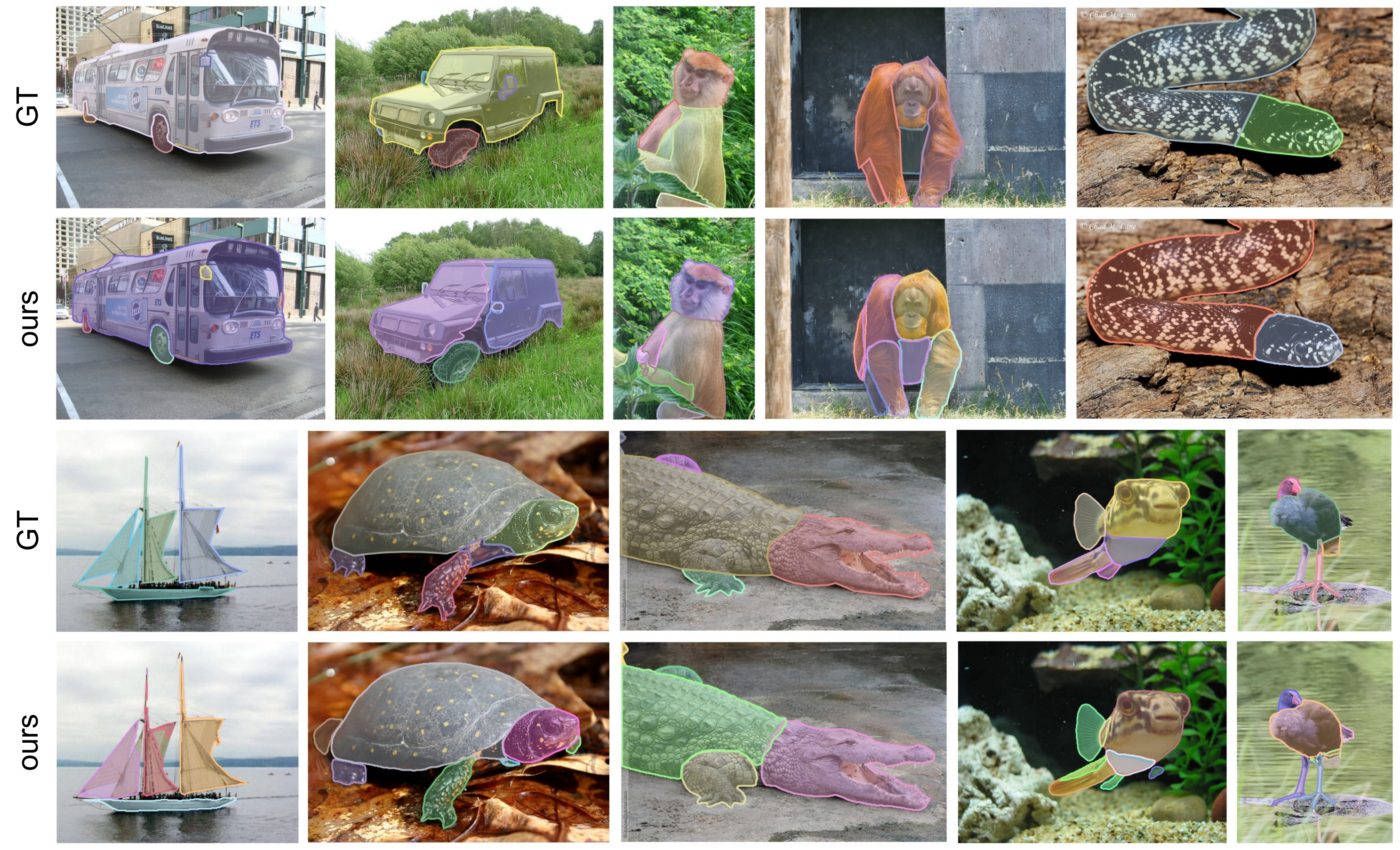}}
    \caption{\small \textbf{Qualitative results on PartImageNet test set.}}
    \label{fig:suppl_qual2}
\end{figure*}

\cref{fig:suppl_qual2} shows the result on PartImageNet test set by our OPS model trained on PartImageNet train and fine-tuned on PartImageNet val. The part predictions perform well on OOD objects and parts. Some of them even discover more reasonable parts than annotations in GT (\eg chimpanzee on $2$nd row and $4$th column). 

\begin{figure*}[ht]
    \centerline{\includegraphics[width=1\linewidth]{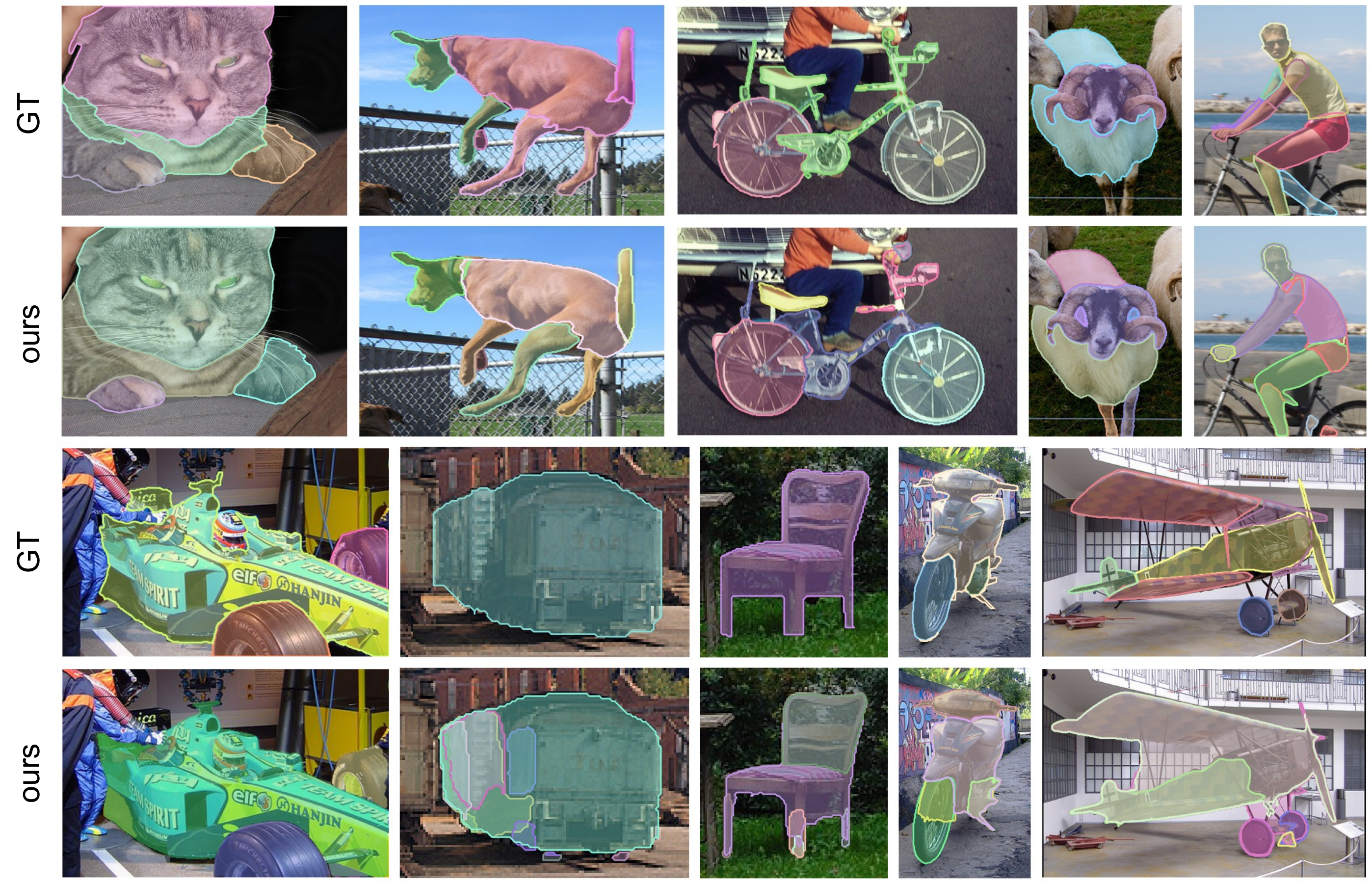}}
    \caption{\small \textbf{Qualitative results on Pascal-Part-58 val set.}}
    \label{fig:suppl_qual3}
\end{figure*}

\cref{fig:suppl_qual3} shows the result on Pascal-Part-58 val set by our OPS model trained on PartImageNet train and fine-tuned on Pascal-Part-58 train. As mentioned in Appendix D, this demonstrates the robustness of OPS to unseen objects and parts in an even more challenging cross-dataset setting. 

